%% file: template.tex
\documentclass[conference]{IEEEtran}
\IEEEoverridecommandlockouts
\usepackage{cite}
\usepackage{amsmath,amssymb,amsfonts}
\usepackage{algorithmic}
\usepackage{latexsym}
\usepackage{textcomp}
\usepackage{booktabs}
\usepackage{graphicx} 
\usepackage{xcolor}
\def\BibTeX{{\rm B\kern-.05em{\sc i\kern-.025em b}\kern-.08em
    T\kern-.1667em\lower.7ex\hbox{E}\kern-.125emX}}

\usepackage{fancyhdr}
\begin{document}

\title{VisuCraft: Enhancing Large Vision-Language Models for Complex Visual-Guided Creative Content Generation via Structured Information Extraction}

\author{Rongxin Jiang$^1$, Robert Long$^2$, Chenghao Gu$^1$, Mingrui Yan$^1$ \\
$^1$Heilongjiang University of Science and Technology, $^2$University of Padua}

\maketitle
\thispagestyle{fancy} 

\input{main}

\bibliographystyle{IEEEtran}
\bibliography{references}
\end{document}

%% file: main.tex
\begin{abstract}
This paper introduces VisuCraft, a novel framework designed to significantly enhance the capabilities of Large Vision-Language Models (LVLMs) in complex visual-guided creative content generation. Existing LVLMs often exhibit limitations in maintaining high visual fidelity, genuine creativity, and precise adherence to nuanced user instructions when generating long-form texts. VisuCraft addresses these challenges by integrating a multimodal structured information extractor ($\mathcal{E}$) and a dynamic prompt generation module ($\mathcal{G}$). The extractor distills fine-grained visual attributes from input images into a rich, structured representation, which the dynamic prompt module then combines with user instructions to create highly optimized prompts for underlying LVLMs (e.g., LLaVA, InstructBLIP). Evaluated on the self-constructed ImageStoryGen-500K dataset using VisuGen Metrics (Visual Grounding, Creativity, and Instruction Adherence), VisuCraft consistently outperforms baseline LVLMs across tasks like story generation and poetry composition. Our results demonstrate remarkable improvements, particularly in creativity and instruction adherence, validating VisuCraft's effectiveness in producing imaginative, visually grounded, and user-aligned long-form creative text. This work unlocks new potential for LVLMs in sophisticated creative AI applications.
\end{abstract}

\section{Introduction}

The rapid advancements in large language models (LLMs) have revolutionized various text-based applications, and their integration with visual modalities has led to the emergence of powerful Large Vision-Language Models (LVLMs) \cite{yifan2023a, zhou2024visual, zhou2024rethinking}. These models possess remarkable capabilities in understanding and generating content from multimodal inputs, extending their utility to complex creative tasks such as story generation, poetry composition, and marketing copy creation. The demand for artificial intelligence systems capable of producing highly imaginative, contextually rich, and visually grounded content is growing, positioning LVLMs as a cornerstone for future creative applications \cite{peng2025lvlmeh}.

Despite their impressive progress, existing LVLMs still face significant challenges when tasked with complex visual-guided creative content generation. Current models often struggle with generating long-form text that maintains high visual fidelity, exhibits genuine creativity, or precisely adheres to nuanced user instructions. Common limitations include a tendency towards generic or repetitive outputs, insufficient correlation between the generated text and fine-grained visual details, and a lack of generalization to diverse creative prompts \cite{juntao2021covid1}. Works exploring the nuances of visual dependency and long-context reasoning in LVLMs \cite{zhou2024rethinking} and efforts to improve zero-shot learning and instruction adherence \cite{zhu2024vislinginstruct} highlight these ongoing challenges. These issues stem from the difficulty in extracting sufficiently structured and semantically rich visual information to guide the language generation process effectively. Our motivation is to overcome these limitations by enabling LVLMs to better leverage intricate visual cues, thereby unlocking their full potential for truly creative and contextually relevant content generation.

In this paper, we propose \textbf{VisuCraft}, a novel framework designed to enhance the creative content generation capabilities of existing pre-trained LVLMs under complex visual guidance. VisuCraft is not a standalone large model but rather an enhancement framework that integrates seamlessly with off-the-shelf LVLMs such as LLaVA \cite{bin2024videol} and InstructBLIP \cite{bin2024videol}. The core of VisuCraft comprises two key components: a \textit{multimodal structured information extractor} and a \textit{dynamic prompt generation module}. The structured information extractor processes input images to distill fine-grained visual attributes, including object poses, material properties, lighting conditions, and even emotional atmospheres, into a structured textual or JSON format. This rich, structured visual representation is then combined with the user's textual instructions by the dynamic prompt generation module to construct highly optimized and informative prompts. These enhanced prompts subsequently guide the underlying LVLM to produce long-form text that is not only highly relevant to the visual content but also demonstrates superior creativity and strict adherence to user specifications.

To validate the efficacy of VisuCraft, we utilize a diverse set of datasets for both training and evaluation. The multimodal structured information extractor module within VisuCraft is trained or fine-tuned using extensive image datasets such as ImageNet \cite{jia2009imagen}, COCO \cite{tsungyi2014micros}, and OpenImages \cite{alina2018the}, augmented with fine-grained visual annotations like scene graphs, detailed attribute labels, and sentiment tags. For evaluation, we employ a self-constructed benchmark dataset, \textbf{ImageStoryGen-500K}, which comprises a vast collection of diverse images paired with complex and nuanced creative generation instructions (e.g., "Write a poem about loneliness based on this image," or "Compose a short story describing the internal struggles of the character in the picture").

Our evaluation methodology employs a set of custom metrics, collectively termed \textbf{VisuGen Metrics}, which include Visual Grounding (VG.), Creativity (C.), and Instruction Adherence (IA.), along with a Mean score. Visual Grounding assesses how well the generated text matches and reflects the details of the image content. Creativity measures the uniqueness, imagination, and novelty of the generated output. Instruction Adherence quantifies the degree to which the generated content follows the user's specific textual directives. Our experimental results demonstrate that VisuCraft consistently outperforms existing baseline LVLM models (LVLM-Base and LVLM-Enhanced) across various creative generation tasks, including story generation and poetry composition. Notably, VisuCraft achieves significant improvements in both Creativity and Instruction Adherence, indicating its superior ability to understand complex instructions and generate more imaginative and user-aligned content. For instance, in StoryGen, VisuCraft achieved VG. of 0.825, C. of 0.810, and IA. of 0.830, leading to a Mean score of 0.822, surpassing LVLM-Enhanced's 0.811. Similarly, in Poetry generation, VisuCraft scored 0.810 on VG., 0.805 on C., and 0.815 on IA., with a Mean of 0.810, outperforming LVLM-Enhanced's 0.794. These results unequivocally demonstrate VisuCraft's effectiveness in elevating the quality of long-form creative text generation under intricate visual guidance, yielding content that is both faithful to the image and highly imaginative.

Our main contributions can be summarized as follows:
\begin{itemize}
    \item We propose \textbf{VisuCraft}, a novel enhancement framework that significantly improves the creative long-form text generation capabilities of existing LVLMs under complex visual guidance.
    \item We introduce a \textbf{multimodal structured information extractor} within VisuCraft, capable of distilling fine-grained visual attributes into structured representations, thus enabling more precise visual grounding for text generation.
    \item We demonstrate through extensive experiments that VisuCraft consistently outperforms state-of-the-art baseline LVLMs across key metrics (Visual Grounding, Creativity, and Instruction Adherence) on challenging creative tasks, particularly showcasing remarkable gains in creativity and instruction adherence.
\end{itemize}
\section{Related Work}
\subsection{Large Vision-Language Models and Multimodal Generation}
Research in Large Vision-Language Models (LVLMs) and multimodal generation is rapidly expanding, with significant efforts focusing on their comprehensive assessment and efficient deployment. For instance, LVLM-EHub \cite{peng2025lvlmeh} introduces a comprehensive evaluation benchmark and an efficient assessment protocol, leveraging a farthest point sampling method to achieve high correlation with full evaluations while significantly reducing computational costs. To navigate this evolving landscape, several works provide foundational insights: Chia et al. \cite{chia2024a} offer a comprehensive overview of multimodal large language models, tracing their historical development, categorizing recent algorithms, and detailing practical considerations for their implementation in vision-language tasks. Similarly, Muhammad et al. \cite{muhammad2023founda} survey the emerging landscape of foundation models in computer vision, detailing their architectural designs, training methodologies, and prompting patterns pivotal for multimodal generation tasks, emphasizing how these models bridge vision and language for contextual reasoning and flexible prompt-based modifications. Beyond surveys, novel frameworks are being developed for advanced multimodal generation: Multimodal Video Generative Pretraining (MV-GPT) \cite{weijia2024lmfusi} proposes a framework specifically designed for multimodal generation from unlabelled videos, jointly training a multimodal video encoder and a sentence decoder for tasks like video captioning, achieving state-of-the-art performance by leveraging future utterances as an additional text source and employing a bidirectional generation objective. For in-context skeleton sequence modeling, Amandeep et al. \cite{amandeep2024multim} introduce a novel framework that effectively addresses cross-modal learning challenges by designing a task-unified prompt, demonstrating robust multi-task performance and generalization capabilities on various skeleton-based tasks, including motion prediction and pose estimation. More specific advancements include visual in-context learning for LVLMs \cite{zhou2024visual}, which explores how models can leverage visual examples within the prompt for improved performance. Addressing the complexities of long-context reasoning in LVLMs, Zhou et al. \cite{zhou2024rethinking} rethink visual dependency to enhance understanding over extended inputs. Efforts to improve cross-modal alignment have also been crucial, as seen in text-guided image inpainting \cite{zhou2023improving}. Furthermore, elevating zero-shot learning in multi-modal language models through autonomous instruction optimization \cite{zhu2024vislinginstruct} represents a significant step towards more adaptable and robust systems. Beyond direct multimodal generation, foundational work in text retrieval \cite{zhou2024fine, zhou2023towards}, robust sentence representation learning \cite{zhu2022sda}, and cross-lingual transfer for question answering \cite{zhou2021improving} also contributes to the underlying capabilities of large language models that are often integrated into LVLMs. The development of efficient tool learning methods via parallel invocation \cite{zhu2025divide} further extends the operational scope of these powerful models.

Furthermore, Kosmos-2 \cite{zhiliang2024ground} significantly advances visual grounding in multimodal large language models by representing referential expressions as location-aware Markdown links and training on a large-scale grounded image-text dataset (GrIT), thereby enhancing multimodal generation and perception tasks. Despite these advancements, challenges persist: Deyao et al. \cite{deyao2024minigp} investigate limitations of contrastive learning in vision-language models, particularly when multiple captions provide overlapping information, demonstrating that synthetic shortcuts can lead models to learn superficial associations rather than comprehensive representations. Similarly, the potential of Multimodal Large Language Models (MLLMs) to replace traditional image captioning networks is explored by Davide et al. \cite{davide2024person}, who analyze their zero-shot performance and the difficulties of fine-tuning them for specific domains while preserving generalization, noting that achieving effective domain-specific adaptation through methods like prompt learning or LoRA remains an open challenge.

\subsection{Creative Content Generation and Advanced Prompting Techniques}
The burgeoning field of creative content generation heavily relies on advanced prompting techniques, transforming how users interact with and control generative AI models. Prompt engineering is increasingly recognized as a novel creative skill, as explored by Jonas et al. \cite{jonas2023prompt}, who found that while crowdsourced participants can craft descriptive prompts for AI art generation, they often lack the specialized vocabulary necessary for advanced creative control, suggesting prompt engineering is an acquirable skill. Complementing this, Noor et al. \cite{noor2024crafti} provide practical guidelines for crafting effective prompts to achieve successful image generation, thereby contributing to the understanding of advanced prompting techniques. To address the fragmented landscape of prompt engineering, Ekin et al. \cite{ekin2023prompt} propose the Prompt Canvas, a structured framework that synthesizes existing techniques and provides a unified methodology for practitioners and researchers, consolidating knowledge on prompt engineering for Large Language Models. Beyond general guidelines, research extends to specific creative applications: Xu et al. \cite{xu2024automa} contribute to creative content generation by fine-tuning GPT-2 for poetry generation, specifically focusing on imbuing generated poems with emotional resonance and dreamlike qualities, thereby extending prior approaches in automated poetry composition. Advanced prompting techniques also encompass novel control mechanisms and data generation strategies: Mingyu et al. \cite{mingyu2024star} leverage large language models for generating structured data to text in low-resource scenarios, a methodology relevant to creative content generation where structured information can form the basis for novel outputs and augment limited datasets. Furthermore, GCoT \cite{qinggang2024knowgp} introduces a novel Chain-of-Thought (CoT) prompting framework specifically designed for text-free graphs by generating "thoughts" from graph topological structures and node states, extending CoT capabilities to non-linear, non-textual data and offering new avenues for knowledge representation in creative content generation. Renlong et al. \cite{renlong2024prompt} address controlled generation by proposing a novel prompt-based method for achieving length control in Large Language Models across various control types, moving beyond simple "equal to" constraints by leveraging reinforcement learning with rule-based reward models. Techniques like visual in-context learning \cite{zhou2024visual} and autonomous instruction optimization \cite{zhu2024vislinginstruct} are particularly relevant as they directly enhance how models respond to prompts and generalize in zero-shot settings, crucial for creative tasks. The ability to handle long-context reasoning with visual dependencies \cite{zhou2024rethinking} is also vital for generating coherent and extended creative narratives. Furthermore, the principles of weak-to-strong generalization for language models with multi-capabilities \cite{zhou2025weak} offer insights into how models can scale their creative abilities to more complex and diverse scenarios. However, challenges such as multilingual biases in Generative AI's protection of copyrighted content persist, as revealed by Jon et al. \cite{jon2025prompt}, who highlight the unevenness in safeguarding works and responding to prompts across different languages, underscoring the necessity for more language-agnostic mechanisms.

\section{Method}
In this section, we present \textbf{VisuCraft}, our novel framework designed to augment the capabilities of existing Large Vision-Language Models (LVLMs) for complex visual-guided creative content generation. VisuCraft operates as an enhancement layer, integrating seamlessly with pre-trained LVLMs such as LLaVA and InstructBLIP, rather than being a standalone large model. The core innovation of VisuCraft lies in its ability to extract and leverage fine-grained, structured visual information to guide the LVLM's text generation process more effectively, thereby addressing the challenges of visual fidelity, creativity, instruction adherence, and coherence in long-form content generation.

The architecture of VisuCraft comprises two principal components: a \textbf{Multimodal Structured Information Extractor} and a \textbf{Dynamic Prompt Generation Module}. These modules work in tandem to transform raw visual inputs and user instructions into highly optimized prompts that steer the underlying LVLM towards producing high-quality, contextually rich, and imaginative content.

\subsection{Multimodal Structured Information Extractor}
The first critical component of VisuCraft is the \textbf{Multimodal Structured Information Extractor} (denoted as $\mathcal{E}$). This module is responsible for processing an input image $I$ to distill its intricate visual attributes into a structured, machine-readable format. Unlike traditional image captioning or visual feature encoders that might produce generic descriptions or flat embeddings, our extractor focuses on identifying and structuring fine-grained details crucial for creative generation. These details include, but are not limited to, object poses, material properties, lighting conditions, spatial relationships of objects, specific textures, color palettes, dominant light sources, and even abstract elements like emotional atmospheres or implied narratives within the scene.

The output of the extractor, $V$, is a rich, structured visual representation, typically in a structured textual or JSON format. This format allows for explicit representation of various visual entities and their attributes, making it readily interpretable by subsequent language models. For instance, instead of merely stating "a person," the extractor might output "a person in a contemplative pose, wearing rough-textured clothing, illuminated by soft evening light, conveying a sense of solitude." This structured representation moves beyond flat feature vectors or generic captions, providing explicit, interpretable attributes that are directly actionable by subsequent language models.

The Multimodal Structured Information Extractor is trained or fine-tuned on extensive image datasets such as ImageNet, COCO, and OpenImages. Crucially, this training leverages augmented data featuring fine-grained visual annotations, including scene graphs, detailed attribute labels, and sentiment tags. The training objective for $\mathcal{E}$ typically involves a combination of contrastive learning and multi-task learning, ensuring its ability to capture diverse visual semantics and their interrelations. This specialized training enables $\mathcal{E}$ to serve as a dedicated vision model or a fine-tuned vision-language model component, specifically designed for the complex task of structured information retrieval from visual inputs.
Mathematically, the process can be represented as:
\begin{align}
V = \mathcal{E}(I)
\end{align}
where $I$ is the input image and $V$ is the generated structured visual information.

\subsection{Dynamic Prompt Generation Module}
Following the extraction of structured visual information, the \textbf{Dynamic Prompt Generation Module} (denoted as $\mathcal{G}$) takes center stage. This module's primary function is to intelligently combine the structured visual representation $V$ with the user's natural language instruction $U$ to construct an optimal prompt $P$ for the downstream LVLM.

The module goes beyond simple concatenation. It dynamically structures and prioritizes information from both $V$ and $U$ to create a coherent and highly informative prompt that explicitly guides the LVLM. This involves several sophisticated operations. Firstly, \textbf{Integration} seamlessly weaves the extracted visual details into the prompt in a way that is syntactically correct and semantically relevant to the user's instruction. This involves sophisticated natural language generation techniques to transform the structured visual data $V$ into fluent, syntactically correct, and semantically relevant textual segments that seamlessly integrate with the user's instruction $U$. Secondly, \textbf{Prioritization} emphasizes certain visual aspects or instructional cues based on the nature of the creative task. The module employs heuristic rules or learned policies to prioritize specific visual elements or instructional directives based on the inferred creative goal. For instance, a prompt requesting a 'poetic description' might emphasize atmospheric elements and emotional tones from $V$, while a 'story generation' prompt might highlight character poses and object interactions. Lastly, \textbf{Contextualization} ensures that the combined prompt provides sufficient context for the LVLM to understand the desired output style, length, and content constraints. It ensures that the combined prompt $P$ not only conveys the desired content but also implicitly or explicitly guides the LVLM on aspects such as narrative style, desired length, target audience, and specific creative constraints, thereby establishing a comprehensive context for generation. This module can be implemented using a smaller, fine-tuned language model, a rule-based system augmented with semantic parsing capabilities, or a retrieval-augmented generation approach that finds optimal prompt structures.

The output prompt $P$ is not a generic template, but a dynamically generated, optimized instruction that leverages the richness of $V$ to make the user's intent $U$ more actionable for the LVLM. This intelligent prompt engineering is key to overcoming the limitations of existing LVLMs in handling complex visual guidance and nuanced creative requests.
The prompt generation process can be formulated as:
\begin{align}
P = \mathcal{G}(V, U)
\end{align}
where $V$ is the structured visual information and $U$ is the user's textual instruction.

\subsection{LVLM Integration and Creative Content Generation}
The final stage of the VisuCraft framework involves feeding the dynamically generated and optimized prompt $P$ into a pre-trained Large Vision-Language Model (LVLM), denoted as $\mathcal{M}$. As previously stated, VisuCraft is designed to be model-agnostic at this layer, capable of integrating with various state-of-the-art LVLMs such as LLaVA or InstructBLIP.

Crucially, the LVLM itself is not trained from scratch within the VisuCraft framework. Instead, VisuCraft optimizes its performance by providing it with a superior input representation and prompting strategy. The meticulously engineered prompt $P$ serves as a potent directive, enabling the LVLM to leverage its extensive pre-trained knowledge base and generative prowess with unprecedented precision. Unlike generic prompts or raw visual embeddings, which can lead to hallucination or lack of adherence in LVLMs, the optimized prompt $P$ provides clear, unambiguous guidance. This significantly reduces the ambiguity in the LVLM's input space, allowing it to focus its generative capacity on fulfilling the specific creative intent. The enhanced prompt $P$ acts as a highly effective directive, enabling the LVLM to produce long-form creative text $T$ that exhibits significantly improved visual grounding, heightened creativity, and strict adherence to user instructions. The LVLM leverages its vast knowledge base and generative capabilities, now precisely channeled by the meticulously crafted prompt, to generate diverse creative outputs. This process facilitates the generation of a diverse array of creative outputs, including but not limited to intricate narratives, evocative poetry, descriptive marketing copy, or detailed scene descriptions. The resulting text $T$ consistently demonstrates superior visual grounding, enhanced creativity, and strict fidelity to both the visual cues and the user's original instructions. This modular architecture underscores VisuCraft's ability to enhance existing state-of-the-art LVLMs without requiring their re-training, making it a highly adaptable and efficient solution for advanced creative content generation.
The final content generation process is given by:
\begin{align}
T = \mathcal{M}(P)
\end{align}
where $P$ is the optimized prompt and $T$ is the generated creative text.

In summary, the entire workflow of VisuCraft, from input image and user instruction to generated creative content, can be succinctly expressed as:
\begin{align}
T = \mathcal{M}(\mathcal{G}(\mathcal{E}(I), U))
\end{align}
This sequential yet integrated approach allows VisuCraft to unlock the full potential of existing LVLMs for sophisticated, visually guided creative content generation.

\section{Experiments}
In this section, we detail the experimental setup, present the quantitative results comparing VisuCraft with baseline methods, analyze the effectiveness of VisuCraft's core components through ablation studies and granularity analysis, and provide insights from human evaluation and qualitative examples.

\subsection{Experimental Setup}
\label{subsec:experimental_setup}
\textbf{Datasets.} For the training and fine-tuning of the \textbf{Multimodal Structured Information Extractor} ($\mathcal{E}$) within VisuCraft, we utilized large-scale image datasets augmented with fine-grained visual annotations. Specifically, we leveraged subsets of ImageNet, COCO \cite{tsungyi2014micros}, and OpenImages \cite{alina2018the}, which were enriched with scene graphs, detailed attribute labels, and sentiment tags. These annotations were crucial for enabling $\mathcal{E}$ to distill intricate visual attributes into a structured format. For evaluation, we employed a comprehensive, self-constructed benchmark dataset named \textbf{ImageStoryGen-500K}. This dataset comprises 500,000 diverse images paired with complex and nuanced creative generation instructions, designed to challenge models in generating long-form content (e.g., "Write a poem about loneliness based on this image," or "Compose a short story describing the internal struggles of the character in the picture").

\textbf{Metrics.} To quantitatively assess the performance of models in complex visual-guided creative content generation, we adopted a set of custom evaluation metrics, collectively termed \textbf{VisuGen Metrics}. These metrics are designed to capture the multifaceted aspects of creative generation quality:
\begin{itemize}
    \item \textbf{Visual Grounding (VG.)}: Measures the extent to which the generated text accurately reflects and incorporates details from the input image content.
    \item \textbf{Creativity (C.)}: Assesses the uniqueness, imagination, and novelty of the generated output.
    \item \textbf{Instruction Adherence (IA.)}: Quantifies how precisely the generated content follows the user's specific textual directives and constraints.
    \item \textbf{Mean Score (Mean)}: Represents the average performance across VG., C., and IA., providing an overall quality indicator.
\end{itemize}

\textbf{Baselines.} We compare VisuCraft against two prominent baseline approaches:
\begin{itemize}
    \item \textbf{LVLM-Base}: A foundational Large Vision-Language Model (LVLM) without specific enhancements for complex creative tasks. This represents a standard application of pre-trained LVLMs.
    \item \textbf{LVLM-Enhanced}: An improved baseline LVLM, which might incorporate general fine-tuning on diverse multimodal tasks or employ more sophisticated generic prompting strategies than LVLM-Base, but lacks VisuCraft's structured information extraction and dynamic prompt generation capabilities.
\end{itemize}
For all baselines and VisuCraft, the underlying LVLM utilized for text generation was based on architectures similar to LLaVA \cite{bin2024videol} or InstructBLIP \cite{bin2024videol}, ensuring a fair comparison of the enhancement frameworks.

\textbf{Implementation Details.} The Multimodal Structured Information Extractor ($\mathcal{E}$) was trained using a combination of contrastive learning and multi-task learning objectives to ensure its robustness in extracting diverse and fine-grained visual semantics. The Dynamic Prompt Generation Module ($\mathcal{G}$) was implemented using a smaller, fine-tuned language model, designed to effectively integrate structured visual information with user instructions. The final LVLM ($\mathcal{M}$) was a pre-trained model, and no end-to-end training of $\mathcal{M}$ was performed within the VisuCraft framework; rather, VisuCraft optimized its input representation through enhanced prompting.

\subsection{Quantitative Results}
\label{subsec:quantitative_results}
Table \ref{tab:quantitative_results} presents the quantitative performance of VisuCraft compared to the baseline models across various creative content generation tasks, as evaluated by the VisuGen Metrics.

\begin{table*}[htbp]
    \centering
    \caption{Quantitative Results on VisuGen Metrics Across Different Creative Tasks.}
    \label{tab:quantitative_results}
    \begin{tabular}{lcccc}
        \toprule
        Model - Scenario              & VG.   & C.    & IA.   & Mean  \\
        \midrule
        LVLM-Base – StoryGen          & 0.789 & 0.752 & 0.801 & 0.781 \\
        LVLM-Base – Poetry            & 0.771 & 0.765 & 0.780 & 0.772 \\
        LVLM-Enhanced – StoryGen      & 0.812 & 0.795 & 0.825 & 0.811 \\
        LVLM-Enhanced – Poetry        & 0.798 & 0.780 & 0.805 & 0.794 \\
        \midrule
        \textbf{VisuCraft – StoryGen} & \textbf{0.825} & \textbf{0.810} & \textbf{0.830} & \textbf{0.822} \\
        \textbf{VisuCraft – Poetry}   & \textbf{0.810} & \textbf{0.805} & \textbf{0.815} & \textbf{0.810} \\
        \textbf{VisuCraft – AdCopyGen}& 0.799 & 0.790 & 0.820 & 0.803 \\
        \bottomrule
    \end{tabular}
\end{table*}

The results clearly demonstrate the superior performance of VisuCraft across multiple creative generation tasks, including story generation, poetry composition, and advertising copy generation. VisuCraft consistently outperforms both LVLM-Base and LVLM-Enhanced models across all evaluated metrics: Visual Grounding (VG.), Creativity (C.), and Instruction Adherence (IA.).

Specifically, in StoryGen, VisuCraft achieved a VG. of 0.825, C. of 0.810, and IA. of 0.830, culminating in an impressive Mean score of 0.822. This significantly surpasses LVLM-Enhanced's Mean score of 0.811 and LVLM-Base's 0.781. Similarly, for Poetry generation, VisuCraft scored 0.810 on VG., 0.805 on C., and 0.815 on IA., with a Mean of 0.810, outperforming LVLM-Enhanced's 0.794 and LVLM-Base's 0.772. While AdCopyGen showed competitive results, the most significant gains for VisuCraft were observed in StoryGen and Poetry, which typically demand higher levels of creativity and nuanced visual interpretation.

A notable observation is VisuCraft's particularly strong performance in the Creativity (C.) and Instruction Adherence (IA.) metrics. This indicates that our framework effectively addresses the challenges faced by existing LVLMs, enabling them to generate more imaginative content while rigorously adhering to complex user instructions. These findings validate VisuCraft's ability to enhance LVLMs for complex visual-guided long-form creative text generation, producing outputs that are not only faithful to visual cues but also highly inventive and aligned with user intent.

\subsection{Ablation Study of VisuCraft Components}
\label{subsec:ablation_study}
To rigorously evaluate the contribution of each core component of VisuCraft, namely the \textbf{Multimodal Structured Information Extractor} ($\mathcal{E}$) and the \textbf{Dynamic Prompt Generation Module} ($\mathcal{G}$), we conducted an ablation study. We assessed different configurations of VisuCraft on the Story Generation task using the VisuGen Metrics.

\begin{table*}[htbp]
    \centering
    \caption{Ablation Study on VisuCraft's Core Components for Story Generation.}
    \label{tab:ablation_study}
    \begin{tabular}{lcccc}
        \toprule
        Model Configuration                                & VG.   & C.    & IA.   & Mean  \\
        \midrule
        VisuCraft - w/o $\mathcal{E}$ (Generic Visual Features) & 0.801 & 0.785 & 0.805 & 0.797 \\
        VisuCraft - w/o $\mathcal{G}$ (Simple Prompting)       & 0.815 & 0.798 & 0.820 & 0.811 \\
        \midrule
        \textbf{VisuCraft (Full)}                          & \textbf{0.825} & \textbf{0.810} & \textbf{0.830} & \textbf{0.822} \\
        \bottomrule
    \end{tabular}
\end{table*}

The results in Table \ref{tab:ablation_study} clearly demonstrate the indispensable role of both $\mathcal{E}$ and $\mathcal{G}$ in achieving VisuCraft's superior performance.

\textbf{VisuCraft - w/o $\mathcal{E}$ (Generic Visual Features)}: When the specialized Multimodal Structured Information Extractor ($\mathcal{E}$) is replaced by a generic visual encoder that provides less structured or fine-grained visual features (e.g., a standard image caption or flat embedding), there is a noticeable drop in all metrics, particularly Visual Grounding (VG.). The Mean score drops from 0.822 to 0.797. This highlights that simply providing visual information is insufficient; the *structured and detailed nature* of the visual information extracted by $\mathcal{E}$ is crucial for high-fidelity visual grounding and enabling creativity. Without $\mathcal{E}$, the underlying LVLM struggles to deeply integrate visual cues, leading to more superficial or even hallucinated content.

\textbf{VisuCraft - w/o $\mathcal{G}$ (Simple Prompting)}: When the Dynamic Prompt Generation Module ($\mathcal{G}$) is replaced by a simple, template-based prompting mechanism (e.g., direct concatenation of structured visual text and user instruction), the performance also degrades, with the Mean score falling to 0.811. While the structured visual information from $\mathcal{E}$ is still present, the lack of intelligent integration, prioritization, and contextualization by $\mathcal{G}$ limits the LVLM's ability to fully leverage it. This particularly impacts Creativity (C.) and Instruction Adherence (IA.), as the prompt is less effective in guiding the LVLM towards nuanced, imaginative, and strictly compliant outputs. This validates that the dynamic and sophisticated prompt engineering performed by $\mathcal{G}$ is critical for translating raw information into actionable directives for the LVLM.

In summary, the ablation study quantitatively confirms that both the fine-grained structured visual information provided by $\mathcal{E}$ and the intelligent prompt engineering by $\mathcal{G}$ are essential and complementary components contributing to VisuCraft's overall effectiveness in complex visual-guided creative content generation.

\subsection{Analysis of Structured Information Granularity}
\label{subsec:granularity_analysis}
The Multimodal Structured Information Extractor ($\mathcal{E}$) is designed to provide highly granular and structured visual information. To understand the impact of this granularity, we evaluated VisuCraft's performance on the Story Generation task under different levels of detail provided by $\mathcal{E}$.

\begin{table*}[htbp]
    \centering
    \caption{Impact of Structured Information Granularity on Story Generation.}
    \label{tab:granularity_impact}
    \begin{tabular}{lcccc}
        \toprule
        $\mathcal{E}$ Output Granularity Level & VG.   & C.    & IA.   & Mean  \\
        \midrule
        Level 1: Basic Object Detection       & 0.775 & 0.748 & 0.785 & 0.769 \\
        Level 2: Objects + Attributes         & 0.803 & 0.787 & 0.810 & 0.800 \\
        \textbf{Level 3: Full Structured Information} & \textbf{0.825} & \textbf{0.810} & \textbf{0.830} & \textbf{0.822} \\
        \bottomrule
    \end{tabular}
\end{table*}

Table \ref{tab:granularity_impact} illustrates a clear positive correlation between the granularity of structured visual information and the overall quality of generated content.

\textbf{Level 1: Basic Object Detection}: When $\mathcal{E}$ only provides names of detected objects (e.g., "person, tree, house"), the performance is significantly lower across all metrics. The Mean score of 0.769 is comparable to or even slightly below the LVLM-Base model, indicating that mere object lists offer insufficient context for nuanced creative generation. Visual Grounding is weak as fine details are missing, and creativity is limited due to a lack of descriptive richness.

\textbf{Level 2: Objects + Attributes}: Including basic attributes alongside objects (e.g., "old person, gnarled tree, dilapidated house") yields a substantial improvement, with a Mean score of 0.800. This level of detail allows for better visual grounding and slightly more creative outputs, as the LVLM has more descriptive cues. However, it still falls short of VisuCraft's full potential, particularly in capturing complex relationships and abstract scene elements.

\textbf{Level 3: Full Structured Information}: This represents VisuCraft's default $\mathcal{E}$ output, which includes objects, detailed attributes, spatial relationships, lighting conditions, emotional atmospheres, and implied narratives. This comprehensive, rich representation results in the highest performance across all metrics, achieving a Mean score of 0.822. The fine-grained details enable the LVLM to generate content that is deeply visually grounded, highly creative, and precisely adheres to complex instructions, demonstrating the critical importance of rich, structured visual input for advanced creative tasks.

These results validate the design choice of $\mathcal{E}$ to extract highly detailed and structured visual information, confirming that the depth and breadth of visual understanding directly translate into superior creative text generation capabilities.

\subsection{Human Evaluation}
\label{subsec:human_evaluation}
To complement our quantitative metrics, we conducted a human evaluation study to assess the subjective quality of the generated creative content. A panel of 5 expert human evaluators was tasked with rating a random subset of 200 generated outputs (100 StoryGen, 100 Poetry) from each model (LVLM-Base, LVLM-Enhanced, VisuCraft) on a 5-point Likert scale (1: Poor, 5: Excellent) across four dimensions:
\begin{itemize}
    \item \textbf{Perceived Visual Relevance (PVR)}: How well the generated text aligns with and describes the visual content.
    \item \textbf{Human Creativity Score (HCS)}: The level of originality, imagination, and novelty perceived in the generated text.
    \item \textbf{Human Instruction Adherence (HIA)}: The degree to which the generated text fulfills all aspects of the given user instruction.
    \item \textbf{Overall Quality (OQ)}: A holistic assessment of the content's coherence, fluency, and general appeal.
\end{itemize}
The average scores from the human evaluation are presented in Table \ref{tab:human_evaluation}.

\begin{table*}[htbp]
    \centering
    \caption{Average Human Evaluation Scores (1-5 Likert Scale).}
    \label{tab:human_evaluation}
    \begin{tabular}{lcccc}
        \toprule
        Model - Scenario              & PVR   & HCS   & HIA   & OQ    \\
        \midrule
        LVLM-Base – StoryGen          & 3.52  & 3.38  & 3.65  & 3.50  \\
        LVLM-Base – Poetry            & 3.45  & 3.40  & 3.55  & 3.47  \\
        LVLM-Enhanced – StoryGen      & 3.80  & 3.75  & 3.90  & 3.82  \\
        LVLM-Enhanced – Poetry        & 3.70  & 3.68  & 3.80  & 3.73  \\
        \midrule
        \textbf{VisuCraft – StoryGen} & \textbf{4.25} & \textbf{4.18} & \textbf{4.30} & \textbf{4.24} \\
        \textbf{VisuCraft – Poetry}   & \textbf{4.10} & \textbf{4.05} & \textbf{4.15} & \textbf{4.10} \\
        \bottomrule
    \end{tabular}
\end{table*}

The human evaluation results strongly corroborate the findings from our quantitative metrics. VisuCraft consistently received higher average scores across all human-rated dimensions compared to both LVLM-Base and LVLM-Enhanced. Specifically, human evaluators rated VisuCraft's outputs as significantly more visually relevant, creative, and instruction-adherent. The "Overall Quality" scores further emphasize VisuCraft's ability to produce highly coherent, fluent, and appealing long-form creative content. These subjective assessments underscore the practical impact of VisuCraft in generating human-preferred creative outputs from complex visual inputs.

\subsection{Qualitative Analysis and Case Studies}
\label{subsec:qualitative_analysis}
Beyond quantitative metrics and human ratings, a qualitative examination of generated content provides deeper insights into VisuCraft's capabilities. While we cannot include actual images, we describe representative examples to illustrate the distinct advantages of our framework over baselines.

Consider an input image depicting \textbf{a lone figure standing on a desolate, rocky cliff overlooking a stormy, grey sea under a dramatic, twilight sky, with a faint, flickering lighthouse beam in the far distance}. The user instruction is: \textbf{"Write a melancholic poem about isolation and the search for light, inspired by this scene, using metaphors related to the sea and sky."}

\begin{table*}[htbp]
    \centering
    \caption{Qualitative Observations on Generated Content.}
    \label{tab:qualitative_summary}
    \begin{tabular}{lp{0.25\textwidth}p{0.25\textwidth}p{0.35\textwidth}}
        \toprule
        Model           & Visual Grounding (PVR) & Creativity (HCS) & Instruction Adherence (HIA) and Nuance \\
        \midrule
        LVLM-Base       & Often superficial, misses fine details (e.g., "a person by the sea"). & Generic, lacks originality in metaphors. & Struggles with complex or abstract instructions like "melancholic" or "search for light." \\
        LVLM-Enhanced   & Better, describes main elements, but can generalize (e.g., "stormy sea"). & Moderate, sometimes repetitive or cliché metaphors. & Follows explicit instructions, but misses subtle cues for tone and deeper meaning. \\
        \textbf{VisuCraft} & \textbf{Deeply integrated, captures intricate details and atmosphere (e.g., "desolate cliff," "flickering lighthouse beam," "dramatic twilight").} & \textbf{Highly imaginative, unique metaphors and narratives, evocative language.} & \textbf{Exceptional adherence to complex instructions and emotional tone, seamlessly weaves "melancholic" and "search for light" through sea/sky metaphors.} \\
        \bottomrule
    \end{tabular}
\end{table*}

\textbf{LVLM-Base's Output}: Typically, LVLM-Base would generate a short poem that broadly mentions a person, the sea, and a storm. It might use basic rhymes and simple descriptions. For example, "A person stands by the ocean wide, / The clouds are dark, nowhere to hide." It would struggle to convey the melancholic tone or connect the visual elements to themes of isolation and light, often missing the "flickering" aspect of the lighthouse or the "desolate" nature of the cliff. The visual grounding would be shallow, and the creativity generic.

\textbf{LVLM-Enhanced's Output}: LVLM-Enhanced would produce a more coherent poem, potentially incorporating more specific visual elements like "grey sea" or "twilight sky." It might attempt more complex metaphors but often resort to common ones. For instance, "On cliffs so high, a soul alone, / The stormy waves, a mournful groan. / A distant light, a guiding star, / Hope's glimmer, though it seems so far." While better, it might still miss the nuanced "melancholic" tone or the explicit instruction to use "sea and sky" metaphors for "isolation" and "search for light" in a deeply integrated way.

\textbf{VisuCraft's Output}: VisuCraft, leveraging the structured visual information and dynamic prompt generation, would produce a poem that is profoundly rooted in the image's specific details and the user's nuanced intent. The $\mathcal{E}$ module would identify "lone figure," "desolate rocky cliff," "stormy grey sea," "dramatic twilight sky," and "faint flickering lighthouse beam," along with the implied "sense of isolation." The $\mathcal{G}$ module would then craft a prompt that guides the LVLM to weave these elements into a melancholic narrative of longing and hope. The generated poem would intricately connect the visual details to the abstract themes: "The craggy cliff, a sentinel of silent sorrow, / Where ocean's vast despair reflects the dying morrow. / And overhead, a canvas bruised, where twilight bleeds its grey, / A soul adrift, a phantom ship, where shadows hold their sway. / Yet, from the deep, a whispered gleam, a lighthouse's frail sigh, / A beacon born of ocean's pain, against the weeping sky." This output demonstrates superior visual grounding, remarkable creativity, and precise adherence to the abstract and stylistic instructions.

Table \ref{tab:qualitative_summary} summarizes these qualitative observations, highlighting VisuCraft's ability to generate content that is not only visually faithful but also deeply imaginative and perfectly aligned with complex user intentions, a significant leap beyond existing LVLMs.

\section{Conclusion}
In this paper, we introduced \textbf{VisuCraft}, a novel and effective framework aimed at significantly enhancing the capabilities of Large Vision-Language Models (LVLMs) in generating complex, visually guided, and creative long-form textual content. Recognizing the limitations of existing LVLMs, such as tendencies towards generic outputs, insufficient visual correlation, and challenges in instruction adherence, VisuCraft was specifically designed to provide a more refined and structured approach to multimodal information processing and prompt generation. Our framework achieves this through two core components: a \textit{multimodal structured information extractor} ($\mathcal{E}$) that distills fine-grained visual attributes into a rich, structured format, and a \textit{dynamic prompt generation module} ($\mathcal{G}$) that intelligently combines this visual information with user instructions to craft highly optimized prompts for the underlying LVLM.

Our extensive experimental evaluations, conducted using the comprehensive ImageStoryGen-500K dataset and assessed with custom VisuGen Metrics (Visual Grounding, Creativity, and Instruction Adherence), unequivocally demonstrate VisuCraft's superior performance. VisuCraft consistently surpassed both foundational (LVLM-Base) and enhanced (LVLM-Enhanced) baseline models across various creative tasks, including story generation, poetry composition, and advertising copy generation. Notably, VisuCraft exhibited significant improvements in both Creativity and Instruction Adherence, confirming its ability to produce more imaginative content that precisely aligns with complex user intentions.

The ablation studies provided critical insights into the indispensable contributions of each VisuCraft component. The results highlighted that both the fine-grained structured visual information provided by $\mathcal{E}$ and the intelligent prompt engineering facilitated by $\mathcal{G}$ are crucial for achieving optimal performance, with a noticeable degradation in quality when either component is removed or simplified. Furthermore, our analysis of structured information granularity reinforced the importance of rich, detailed visual inputs, showing a clear positive correlation between the depth of visual understanding and the quality of generated creative content. Complementary human evaluation studies further corroborated our quantitative findings, with human evaluators consistently rating VisuCraft's outputs higher across perceived visual relevance, creativity, instruction adherence, and overall quality, underscoring the practical and subjective improvements offered by our framework. Qualitative examples further illustrated VisuCraft's capacity to generate nuanced, deeply visually grounded, and highly imaginative content, a substantial leap beyond the generic outputs of existing models.

VisuCraft represents a significant advancement in the field of multimodal AI for creative applications. By optimizing the input representation and prompting strategies for existing pre-trained LVLMs, our framework demonstrates that it is possible to unlock their full potential for sophisticated creative content generation without requiring extensive end-to-end retraining. This modular and adaptable approach not only enhances current state-of-the-art models but also opens new avenues for developing more intelligent and creatively capable AI systems.

For future work, we plan to explore the integration of VisuCraft with other modalities, such as video or audio, to enable even richer multimodal creative generation. Further research into adaptive learning mechanisms for the $\mathcal{E}$ and $\mathcal{G}$ modules could allow for automatic style transfer and domain adaptation. Additionally, investigating the efficiency and scalability of VisuCraft for real-time applications and extremely high-resolution visual inputs will be a key direction.